\documentclass[10pt,twocolumn,letterpaper]{article}

\usepackage{wacv}
\usepackage{times}
\usepackage{epsfig}
\usepackage{graphicx}
\usepackage{amsmath}
\usepackage{amssymb}
\usepackage{algorithm}
\usepackage[noend]{algpseudocode}

\wacvfinalcopy 

\def\Mat#1{{\boldsymbol{#1}}}
\def\wacvPaperID{454} 

\ifwacvfinal\fi
\begin{document}

\title{3DCapsule: Extending the Capsule Architecture to Classify 3D Point Clouds}

\author{Ali Cheraghian \\
Australian National University, Data61-CSIRO\\
{\tt\small ali.cheraghian@anu.edu.au}
\and
Lars Petersson \\
Data61-CSIRO\\
{\tt\small lars.petersson@data61.csiro.au}
}

\maketitle
\ifwacvfinal\thispagestyle{empty}\fi

\begin{abstract}
This paper introduces the \textbf{3DCapsule}, which is a 3D extension of the recently introduced Capsule concept that makes it applicable to unordered point sets. The original Capsule relies on the existence of a spatial relationship between the elements in the feature map it is presented with, whereas in point permutation invariant formulations of 3D point set classification methods, such relationships are typically lost. Here, a new layer called \textbf{ComposeCaps} is introduced that, in lieu of a spatially relevant feature mapping, learns a new mapping that can be exploited by the 3DCapsule. Previous works in the 3D point set classification domain have focused on other parts of the architecture, whereas instead, the 3DCapsule is a drop-in replacement of the commonly used fully connected classifier. It is demonstrated via an ablation study, that when the 3DCapsule is applied to recent 3D point set classification architectures, it consistently shows an improvement, in particular when subjected to noisy data. Similarly, the ComposeCaps layer is evaluated and demonstrates an improvement over the baseline. In an apples-to-apples comparison against state-of-the-art methods, again, better performance is demonstrated by the 3DCapsule.
\end{abstract}

\section{Introduction}

3D object classification of point clouds is an important task as laser scanners, or other depth sensors, generating point clouds are now a commodity on, e.g.,  autonomous vehicles, surveying vehicles, service robots and drones. There has been less progress using deep learning methods in the area of point clouds compared to 2D images and videos, partly because the data in a point cloud are typically unordered as opposed to the pixels in a 2D image, which means standard deep learning architectures are not applicable. Nevertheless, some works in the area use architectures circumventing this issue.



\begin{figure*}
\centering
\includegraphics[height=2.3cm]{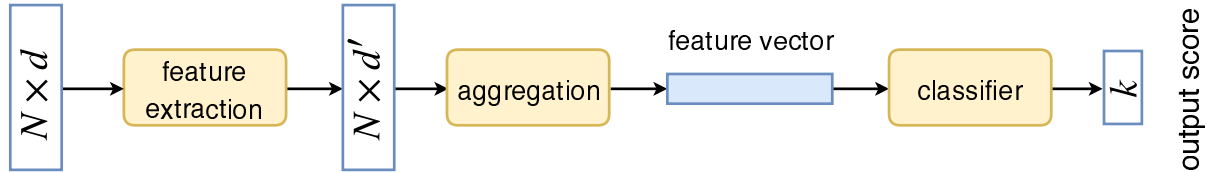}
\caption{A typical architecture of a 3D point cloud classification system. The system takes $N$ points of dimension $d$ and extracts point features of dimension $d^{\prime}$, followed by an aggregation module that is tasked with building a feature vector invariant to point permutation. Lastly, a classifier is used to classify the resulting feature vector into one of the $k$ classes.}
\label{fig:typical_architecture}
\end{figure*}

An architecture for classification that works on raw, unordered, 3D point cloud data often contains three key modules: a feature extraction layer for obtaining a richer representation of input point sets, an aggregation layer acting as a symmetric function aggregating information from points as well as removing the effect of varying ordering (point permutation), and lastly a classifier layer to classify the entire input point set (Figure~\ref{fig:typical_architecture}). The pioneer work in this area is PointNet~\cite{Article1}, which uses a multilayer perceptron (mlp)\cite{Article42} as a feature extraction layer, a max pooling operation as a symmetric function for aggregation, and a fully connected (fc) perceptron as a classifier. 

More recent methods improve upon the first two stages, feature extraction and aggregation~\cite{Article2,Article24,Article27,Article37,Article29}. Feature extraction is improved by getting a richer description using local information instead of only global information~\cite{Article2,Article24,Article27}. In~\cite{Article37}, aggregation is improved by using NetVlad \cite{Article22}. In the context of 3D point classification, there has, however, not been much effort improving the last part, the actual classification layer.


In this paper, we propose a novel extension of Capsule Networks~\cite{Article33} for the purpose of classification of 3D point clouds. The capsule concept was first proposed in~\cite{Article35} and later extended using a dynamic routing algorithm~\cite{Article33}. To date, the capsule concept has not yet found widespread use as it is a recent innovation. However, where it has been used it has indeed improved the overall classification performance, for example,~\cite{Article33} improves upon state-of-the-art on the long time saturated MNIST dataset~\cite{Article36}. The capsule routing algorithm learns global coherence by enforcing part-whole relationships to be learned. For example, if a person's lip is replaced by an eye in a face, using the routing capsule algorithm chances of classifying the image as a face are reduced. Thanks to this capability, the capsule concept has some structural advantages over conventional networks which typically are only capable of extracting local translation invariances. 

In this paper, we extend the Capsule Network concept to be applicable to 3D point cloud data, and we investigate the utility of such a structure when inserted in recent popular methods for 3D point cloud classification. To the best of our knowledge, this is the first time that the capsule concept has been applied to 3D point cloud data. Specifically, we investigate the effect on classification performance as well as robustness to noise via an ablation study considering alternative algorithms for the various parts. This new network design, called 3DCapsule, is non-trivial as more powerful aggregation functions used to introduce an ordering of the data, such as NetVlad, may not preserve a spatial relationship between the data points which the intuition behind the original Capsule Network design relies on.

The main contributions of our work are as follows:
\begin{itemize}
   \item We present a novel classifier, \textbf{3DCapsule}, applicable to 3D point cloud classification. It is a drop-in replacement of the typically used fully connected classifier. 
   \item We add a new layer, \textbf{ComposeCaps}, which in lieu of lost spatial relationships caused by permutation invariance, learns a new, useful, mapping of capsules that can be exploited by the capsule network. 
   \item We demonstrate the benefit of our 3DCapsule and ComposeCaps via ablation studies against baseline architectures. We investigate both classification performance as well as robustness to outlier points and point perturbation. 
   
\end{itemize}

\section{Related Work}

There are three common categories of 3D object classification based on deep learning networks. The first category uses a volumetric representation of the data as input to the network~\cite{Article10,Article11}. The second category is a view-based method~\cite{Article13,Article14}, which projects the 3D shape onto a 2D image such that more classical 2D approaches can be utilized. The third category uses raw point cloud data as input which the network then operates on directly~\cite{Article2,Article24,Article27,Article28,Article29}. In this work, we focus on 3D object classification based on the latter, raw point cloud representation. This is because it does not suffer from severe scalability issues like the volumetric representation does, and it does not make any {\it a priori} assumptions onto which 2D planes, and how many, that the point cloud should be projected on like the view-based methods do.


Methods operating directly on raw point cloud data typically contains three main components: {\it feature extraction}, {\it aggregation}, and a {\it classifier} (see Figure~\ref{fig:typical_architecture}). PointNet~\cite{Article1} is the first work which consumes raw point clouds directly without using any volumetric or view-based data representations.  PointNet uses a multi-layer perceptron (mlp) to extract features from point sets, and instances of Spatial Transformer Networks (STN)~\cite{Jaderberg:2015:STN:2969442.2969465} in order to make it robust to certain 
transformations. Furthermore, a max pooling layer removes permutation, and aggregates information from the points. Classification is done via a fully connected layer at the end.

Recently, a few methods have been introduced which improve the performance of the feature extraction module of PointNet \cite{Article1}. PointNet++~\cite{Article2} extracts richer features which also consider local information. A hierarchical neural network is proposed which applies PointNet \cite{Article1} recursively on partitions of point sets giving it the ability to learn local features from a range of contextual scales. \cite{Article24}, on the other hand, proposes an Edge Convolution (EdgeConv) operation in order to capture local structures by constructing a local neighborhood graph and applying convolution-like operations on the edges connecting neighboring pairs of points. \cite{Article27} introduces a permutation invariant network, called Self-Organization (SO-Net) network, which utilizes a spatial distribution of point clouds. SO-Net also uses a point cloud autoencoder as pre-training to enhance the performance.

Some methods, in turn, instead improve the aggregation module of PointNet \cite{Article1}. \cite{Article29} proposes a structure similar to PointNet++ \cite{Article2} but replaces the mini-pointnet with a spectral convolution operation. To achieve this, a new pooling method, called Recursive cluster pooling is introduced. This pooling strategy aggregates information in the spectral domain from clusters that are related to each other.  \cite{Article37} proposes a new pooling method based on the NetVlad~\cite{Article22} idea. There, unordered point sets are mapped, based on clustering, to a new space that is invariant to permutation enabling a richer feature representation, than, e.g., max pooling.

\begin{figure*}
\centering
\includegraphics[height=4cm]{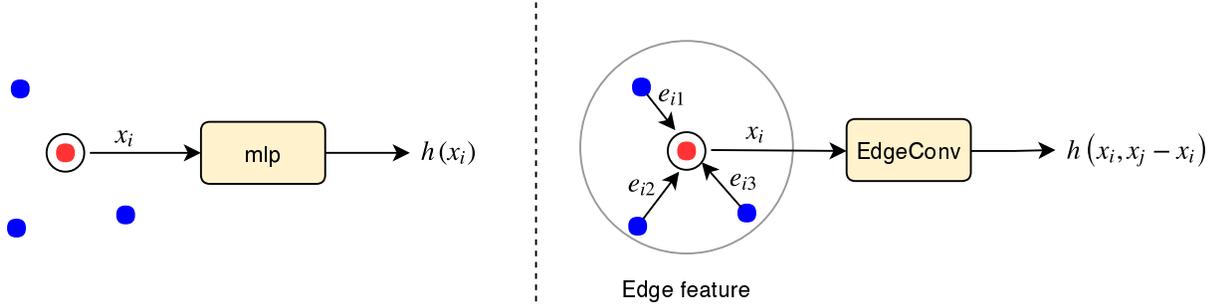}
\caption{Two examples of feature extraction: PointNet \cite{Article1} (left) and EdgeConv (right). In PointNet \cite{Article1}, the output of the feature extraction module $h(\textbf{x}_{i})$ is only related to the point itself, whereas the EdgeConv output $h(\textbf{x}_{i},\textbf{x}_{j}-\textbf{x}_{i})$ depends on the point itself as well as points in its neighborhood.}
\label{fig:feature_extraction}
\end{figure*}

To the best of our knowledge, classification of the representations in these architectures has been limited to using a fully connected layer. Here, we propose an alternative, 3DCapsule, an extension of Capsule Networks~\cite{Article33}. Capsules are originally introduced as an alternative to convolutional neural network (CNN), which is the most common network in deep learning. Capsules can learn a representation of an image that is more robust to spatial and pose variations than a typical CNN thanks to using a vector output instead of a scalar one, and by replacing the max-pooling operation with a dynamic routing algorithm.
Although Capsules have some advantages over CNNs, as demonstrated by the encouraging results on 2D image classification~\cite{Article33,Article38} and segmentation \cite{Article41}, there is currently no work proposed based on the Capsule Network for 3D object classification. Therefore, in this paper, we propose an extension of Capsule Networks such that it becomes applicable to 3D point cloud data, and investigate the utility for the purpose of 3D object classification.

\section{Method}


In this section, we describe our contribution named \textbf{3DCapsule}, which is an essential building block in our proposed method. Putting 3DCapsule into context, we first describe briefly a typical 3D object classification architecture operating on point sets from point clouds. This includes a feature extraction stage and an aggregation stage, the latter which should be invariant to any permutation of the points in the point set. Once these typical elements are explained, we move on to describing our proposed method, 3DCapsule, in detail.

\subsection{Feature extraction and aggregation}

Suppose an unordered point set ${\textbf{X}}={\left \{ {\textbf{x}_{1}},{\textbf{x}_{2}},...,{\textbf{x}_{n}} \right \}}$ where $\textbf{x}_{i}\in\mathbb{R}^{d}$, a set function needs to be defined that is invariant to any permutation of the point set,
\begin{align*}
f(\textbf{x}_{1},\textbf{x}_{2},...,\textbf{x}_{n}) \approx  g(h(\textbf{x}_{1},\theta  )),h(\textbf{x}_{2},\theta  )),...,h(\textbf{x}_{n},\theta  ))
\end{align*}
where $f$ is the set function, $h$ is the feature extraction function, $g$ is the aggregation function which needs to be invariant to permutation of points, and $\theta$ represents a set of arguments associated with $\textbf{x}_i$.  

The feature extraction module $h(\textbf{x}_{i},\theta )$ maps raw point sets to a new, in most cases high dimensional, space where each point (or region around a point) has a richer representation. One choice of $h(\textbf{x}_{i},\theta )$, which is used in PointNet \cite{Article1}, is $h(\textbf{x}_{i}) : \mathbb{R}^{d}\rightarrow\mathbb{R}^{d^{\prime}},\theta =\left \{ \emptyset \right \} $. It is there implemented using a shared mlp layer, where the mlp is applied separately to each point without considering any local neighborhood. This results in extracting features carrying global information only. Another choice of $h(\textbf{x}_{i},\theta )$, which is proposed by~\cite{Article24} called EdgeConv, is $h(\textbf{x}_{i},\textbf{x}_{j}-\textbf{x}_{i}) : \mathbb{R}^{d}\times \mathbb{R}^{d}\rightarrow\mathbb{R}^{d^{\prime}},\theta =\left \{ \textbf{x}_{j}-\textbf{x}_{i} \right \} $. In this case, point sets are represented by a directed graph and edge features based on $k$-nearest neighbors are calculated. EdgeConv~\cite{Article24} is applied to edges extracting features which encode both global {\it and} local information from each point based on its neighborhood (see Figure~\ref{fig:feature_extraction}).

\begin{figure}
\centering
\includegraphics[height=4.5cm]{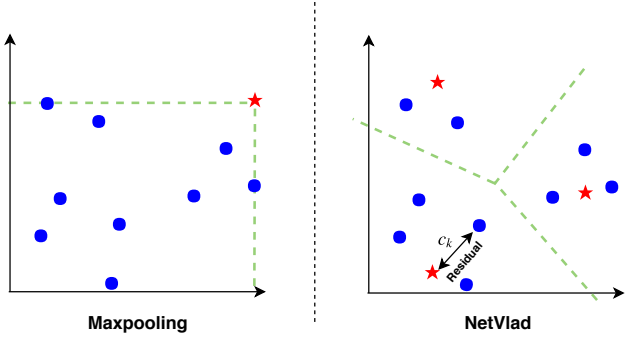}
\caption{Two examples of aggregation methods: Maxpooling (left) and NetVlad \cite{Article22} (right) in a two dimensional space. In Maxpooling, the maximum point is selected in each dimension, so it derives a point which is invariant to permutation. NetVlad, based on predefined centers, computes the residuals between each center and all the points in its neighborhood rendering an invariant representation of the point set.}
\label{fig:aggregation}
\end{figure}

Point sets resulting from a typical 3D scanning device, such as a LIDAR, are inherently unordered, although their {\it spatial distribution} is still invariant to permutation. Hence, we need a function that can aggregate per-point features into a feature vector that is independent of any ordering of the points in the point set. A simple and effective choice of this function, $g$, is the max pooling operation, which is used in PointNet \cite{Article1} and is invariant to permutation. It is also possible to use aggregation methods based on clustering, such as NetVlad~\cite{Article22}. In NetVlad, a few initial point centers are selected via k-means~\cite{Article43} clustering and residual distances between each center and its neighborhood points are calculated. As proven in~\cite{Article37}, this renders a representation invariant to permutation (see Figure~\ref{fig:aggregation}).

Finally, via a collection of $h(\textbf{x}_{i},\theta )$, corresponding values of $f$ can be computed to form a vector $\textbf{f}=[f_{1},f_{2},...,f_{m}]^{T}$ where $\textbf{f}\in \mathbb R^{m}$, which represents a feature vector of an input point set. This feature vector is invariant to any permutation of the input point set.

\subsection{3DCapsule}
\begin{figure*}
\centering
\includegraphics[height=5.5cm]{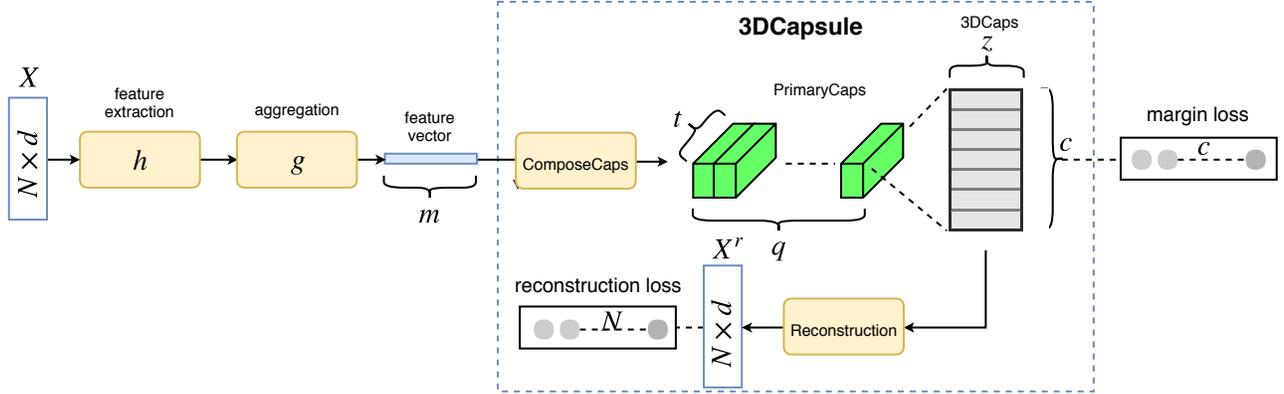}
\caption{The 3DCapsule Architecture. The network takes $N$ points of dimension $d$. Feature extraction, $h$, and aggregation, $g$, create the feature vector $\textbf{f}$ which is input to the 3DCapsule. The 3DCapsule has three main parts: ComposeCaps maps the feature vector to PrimaryCaps which, in turn, creates $q$ primaryCaps of dimension $t$. The routing-by-agreement algorithm is used to couple the PrimaryCaps to the $c$ 3DCaps, each of dimension $z$. The reconstruction block attempts to reconstruct the input data and has a corresponding reconstruction loss.}
\label{fig:3DCapsule_architecture}
\end{figure*}

Here, we introduce our contribution called \textbf{3DCapsule}. The derivation will follow the notation shown in Figure~\ref{fig:3DCapsule_architecture} and~\cite{Article33}.

Suppose we have $q$ primary capsules where each one has dimension $t$ and $c$ 3DCaps each of dimension $z$. \textbf{ComposeCaps} is, here, a fully connected layer mapping the feature vector to primary capsules:  

\begin{align}
\textbf{P}=\phi(\Mat{\left ( W^{C} \right)}^{T}.\textbf{f}+\textbf{b}^{C})
\end{align}
where $\Mat{W^{C}}\in\mathbb R^{m\times t.q}$ is the weight matrix of ComposeCaps, $\textbf{b}^{C}\in \mathbb R^{t.q}$ is its bias, $\phi$ is a sigmoid activation function, and $\textbf{P}\in \mathbb R^{t.q}$ contains the collection of all, newly composed, primary capsules as a vector.

The new capsules are then retrieved by splitting $\textbf{P}$ as follows,
\begin{align}
\textbf{p}_{i}=\left [P_{(i-1)t+1},...,P_{i.t}  \right ]^{T}, i =1,..,q
\end{align}
where $\textbf{p}_{i}\in \mathbb R^{t}$ is a capsule at the primary capsule layer.

A squashing function is defined which maps capsule values to the range $[0,1]$,

\begin{align}
\textbf{v}=squash(\textbf{s})=\frac{\left \| \textbf{s} \right \|^{2}}{1+\left \| \textbf{s} \right \|^{2}}\frac{\textbf{s}}{\left \| \textbf{s} \right \|^{2}}
\label{eq:squash_fcn}
\end{align}
where $\textbf{v}$ is the output vector of a capsule and $\textbf{s}$ is its input. So, in order to find the output of each primary capsule the above squashing function is applied,
\begin{align}
\textbf{u}_{i}=squash(\textbf{p}_{i})\label{eq:2}
\end{align}
where $\textbf{u}_{i}\in \mathbb R^{t}$ is the output vector of a primary capsule.

The prediction vectors $\hat{\textbf{u}}_{ij}$ are obtained by 
\begin{align}
\hat{\textbf{u}}_{ij}=\Mat{W}^{p}_{ji}\textbf{u}_{i}
\end{align}
where $\Mat{W}^{p}_{ji} \in \mathbb R^{c\times t}$, and the input to the 3DCaps layer is a weighted sum over all prediction vectors $\hat{\textbf{u}}_{ij}$ from the capsules in the primary capsule layer
\begin{align}
\textbf{s}_{j}=\sum _{i}c_{ij}\hat{\textbf{u}}_{ij}
\end{align}
where $c_{ij}$ are coupling coefficients that are determined by the routing algorithm in an iterative loop
\begin{align}
c_{ij}=\frac{exp(y_{ij}))}{\sum _{k}exp(y _{ik}))}
\label{eq:c_softmax}
\end{align}
where the initial logits $y_{ij}$ are the log prior probabilities that capsule $i$ should be coupled to capsule $j$. 



Finally, the output of 3DCaps is determined by the same routing algorithm as in the original Capsule formulation~\cite{Article33}(see Algorithm~\ref{alg:routing}).

\makeatletter
\def\BState{\State\hskip-\ALG@thistlm}
\makeatother

\begin{algorithm}
\caption{Routing Algorithm}\label{euclid}
\begin{algorithmic}[1]
\Procedure{Routing($\hat{\textbf{u}}_{ji}$,$r$)}{}
\State  for all capsule $i$ in PrimaryCaps layer and capsule $j$ in 3DCaps layer: $y_{ij}\gets 0$
\For{\textit{$r$ iterations}}
        \State \textit{for all capsule $i$ in PrimaryCaps layer: \\ \qquad \qquad $c_{i} \gets softmax(y_{i}$)}, eq~\ref{eq:c_softmax}
        \State \textit{for all capsule $j$ in 3DCaps layer:   \\ \qquad \qquad $\textbf{s}_{j} \gets \sum _{i}c_{ij}\hat{\textbf{u}}_{ij}$)}
        \State \textit{for all capsule $j$ in 3DCaps layer:  \\ \qquad \qquad $\textbf{v}_{j} \gets squash(\textbf{s}_{j})$)}, eq~\ref{eq:squash_fcn}
        \State \textit{for all capsule $i$ in PrimaryCaps and capsule $j$ \\ \qquad \qquad in 3DCaps layer: $y_{ij} \gets y_{ij}+\hat{\textbf{u}}_{ij}.\textbf{v}_{j}$}
        
\EndFor
\Return $\textbf{v}_{j}$

\EndProcedure
\end{algorithmic}
\label{alg:routing}
\end{algorithm}

The loss function used in the 3DCapsule is the sum of the margin loss and the reconstruction loss, 
\begin{align}
L = \sum_{j=1}^{c}L_{j}+\alpha \sum_{i=1}^{N}\|x^{r}_{i}-x_{i}\|^{2}
\end{align}
\begin{align}
\begin{split}
L_{j}&=T_{j}max(0,m^{+}-\left \| v_{j} \right \|)^2\\
&+\lambda (1-T_{j})max(0,\left \| v_{j} \right \|-m^{-})^2
\end{split}
\end{align}

where $T_{j}$ is equal to 1 if class $j$ is correctly predicted, otherwise 0. The parameters were set as follows: $m^{+}=0.9$, $m^{-}=0.1$, and $\lambda = 0.5$. The total margin loss is calculated by summing the individual margin loss for all $c$ classes. $N$ is the number of points in the set, $x^{r}_{i}$ is the reconstruction of point $x_{i}$, and $\alpha=0.0005$.

\section{Experiments}

In this section, we provide an ablation study to investigate the effect of the 3DCapsule using different feature extractors and aggregation modules. The investigation includes accuracy measures as well as robustness with respect to corruptions such as outlier points and Gaussian perturbation noise. We also evaluate the effect of ComposeCaps and reconstruction loss within the 3DCapsule.

\subsection{Network Architecture and Training}

In all our experiments, we used the following parameters during network training. The Adam optimizer~\cite{Article40} was used with an initial learning rate of 0.001, which was divided by 2 every 20 epochs, and a batch size of 16.
Relu and Batch Normalization(BN)~\cite{Article45} were also used, except in the 3DCapsule part. The network was implemented using TensorFlow~\cite{Article46} and executed on an NVIDIA GTX1080Ti.    

We used feature extraction modules from two different architectures (PointNet \cite{Article1} and EdgeConv \cite{Article24}). 
In the case of the PointNet structure, we employed a spatial transformer network of size $3\times 3$, followed by two shared mlp layers (64,64), followed by the second spatial transformer network of size $64\times 64$, ending with three mlp layers (64,128,1024). 
For the EdgeConv-based feature extraction module, the first EdgeConv block used three shared mlp layers (64,64,64) and the second EdgeConv consisted of a shared mlp of size (128). 
Finally a shared mlp of size (1024) was used to concatenate all features together. 

In terms of aggregation modules, we also used two different approaches (max pooling and NetVlad \cite{Article22}). In the case of max pooling, we got a feature vector of size 1024. Using NetVlad to retrieve a point feature vector, the number of cluster centers were set to 128 and the last mlp layer was set to size 128. In order to be compatible with NetVlad, the last mlp layer of the PointNet  and EdgeConv feature extractors were replaced with an mlp layer of size 128, creating a feature vector output of size $128\times 128$ in the case of NetVlad.

Finally, as the classifier, we again used two different approaches (fully connected classifier as the baseline, and our contribution \textbf{3DCapsule}. In the case of the fully connected classifier, we used three fully connected layers of sizes (512,256,\#classes), with a dropout~\cite{Article47} of $0.7$ in the first two layers. When investigating our 3DCapsule, we used a fully connected layer of size 4000 as our \textbf{ComposeCaps} layer to map the feature vector output to the primary capsule input. 500 primary capsules with a dimension of 8 was used. In the 3DCaps layer, each capsule was of dimension 4.     

\subsection{Datasets}
In our experiments, we used two versions of the ModelNet dataset~\cite{Article10}, ModelNet10 and ModelNet40. The ModelNet40 contains $12,311$ CAD models from 40 different classes, which were divided into $9,843$ training samples and $2,468$ testing samples. The ModelNet10 was split into $3,991$ training samples and $908$ testing samples. We followed the same experimental settings as in~\cite{Article1}. In order to have a fair comparison, the prepared ModelNet 10/40 dataset from \cite{Article2} is used, where each model is represented by $10,000$ points. 

\begin{table}[t]
\caption{Ablation study on ModelNet40. We evaluate the effect of the 3DCapsule in comparison to the fully connected classifier baseline with various combinations of feature extraction and aggregation modules.}
\label{table:ablation_study}
\begin{center}
\scalebox{.85}{

\begin{tabular}{lccc}
\hline\noalign{\smallskip}
feature extraction & aggregation&classifier & Accuracy\\
\noalign{\smallskip}
\hline
\noalign{\smallskip}
PointNet \cite{Article1} &Maxpooling& FC  & 89.2\\

PointNet \cite{Article1} &Maxpooling& 3DCapsule  & \textbf{89.9}\\
\hline\hline
PointNet \cite{Article1}  &NetVlad \cite{Article22}& FC  & 87.2\\
PointNet \cite{Article1}  &NetVlad \cite{Article22}& 3DCapsule  & \textbf{90.6}\\
\hline\hline
EdgeConv \cite{Article24} &Maxpooling&FC& 92.2 \\
EdgeConv \cite{Article24} &Maxpooling&3DCapsule& \textbf{92.4} \\
\hline\hline
EdgeConv \cite{Article24}  &NetVlad \cite{Article22}&FC & 91.2 \\

EdgeConv \cite{Article24}  &NetVlad \cite{Article22}&3DCapsule & \textbf{92.7} \\

\hline
\end{tabular}}
\end{center}
\end{table}

\begin{table}
\centering
\label{table:ablation_studyPointNet}
\caption{Ablation study on ModelNet40. We evaluate the effect of 3DCapsule on PointNet++\cite{Article2} with various number of points representation}
\vspace{3mm}
\label{table:PointNet}
\scalebox{.9}{

\begin{tabular}{lccc}
\hline
method & Classiifer & Representation & Accuracy \\ \hline
PointNet++ \cite{Article2} & FC & $1024 \times 3$ & 90.7 \\ 
PointNet++ \cite{Article2} & 3DCapsule & $1024\times3$ & \textbf{91.5} \\ \hline
\hline
PointNet++ \cite{Article2} & FC & $5000\times6$ & 91.9 \\ 
PointNet++ \cite{Article2} & 3DCapsule & $5000\times6$ & \textbf{92.4} \\ \hline
\end{tabular}}
\end{table}

\subsection{Ablation study of the 3DCapsule}
\label{sec:ablation_study}

In this section, we evaluate the effect of the 3DCapsule with respect to different feature extractors and aggregation modules. The baseline classifier used to compare with, is a fully connected layer (fc) as that is what is employed in comparable works operating on raw point clouds. Furthermore, we use two feature extraction modules based on PointNet \cite{Article1} and EdgeConv\cite{Article24}, and two aggregation approaches, max pooling and NetVlad\cite{Article22}.

From Table~\ref{table:ablation_study}, it is clear that regardless of which feature extraction method and/or aggregation method that is utilized, our 3DCapsule improves over using a regular fully connected layer as a classifier. The biggest improvements over using a fully connected classifier is observed when combined with the aggregator NetVlad. This is natural, as compared to max pooling, NetVlad outputs a much richer representation of the data. A fully connected classifier may not be able to make use of this richer information, whereas our 3DCapsule can due to the fact that it takes advantage of the routing algorithm and capsule vector representation.

In Table~\ref{table:PointNet}, we demonstrate that our method also benefits PointNet++~\cite{Article2}, which further shows the applicability to commonly used point cloud classification architectures. PointNet++ is capable of taking advantage of a richer point representation with additional, normal, features, $5000 \times 6$.   

\subsection{Effect of ComposeCaps}
In this section, we evaluate the effect of ComposeCaps in our proposed method. We, again, test the effect using the two feature extraction operators PointNet \cite{Article1} and EdgeConv \cite{Article24} as well as the two aggregation methods, Maxpooling and NetVlad \cite{Article22}. The result is shown in Table~\ref{table:effect_composecaps}. It can be observed that the ComposeCaps layer does indeed improve the performance in all combinations, demonstrating the utility of ComposeCaps.    

\begin{table}[t]
\caption{The effect of the capsule mapping layer \textbf{ComposeCaps} inside the 3DCapsule.}
\label{table:effect_composecaps}
\begin{center}
\scalebox{.85}{

\begin{tabular}{lcc}
\hline\noalign{\smallskip}
Method & ComposeCaps & Accuracy\\
\noalign{\smallskip}
\hline
\noalign{\smallskip}
PointNet \cite{Article1}+Maxpooling &No  & 89.4\\
PointNet \cite{Article1}+Maxpooling  &Yes  & \textbf{89.9}\\
\hline
\hline
PointNet \cite{Article1}+NetVlad \cite{Article22} &No  & 89.9\\
PointNet \cite{Article1}+NetVlad \cite{Article22}  &Yes  & \textbf{90.6}\\
\hline
\hline
EdgeConv \cite{Article24}+Maxpooling &No&92.3 \\
EdgeConv \cite{Article24}+Maxpooling  &Yes& \textbf{92.4} \\
\hline
\hline
EdgeConv \cite{Article24}+NetVLad \cite{Article22} &No& 92.4 \\
EdgeConv \cite{Article24}+NetVlad \cite{Article22}  &Yes& \textbf{92.7} \\
\hline
\end{tabular}}
\end{center}
\end{table}

\subsection{Effect of Reconstruction Loss}
In this section, we evaluate the effect of reconstruction loss in the 3DCapsule. As shown in Table~\ref{table:effect_reconstruct}, the reconstruction loss is crucial in all structures used in this paper.

\begin{table}
\centering
\label{table:effect_reconstruct}

\caption{The effect of the reconstruction loss function inside the 3DCapsule.}
\vspace{3mm}
\scalebox{.61}{
\begin{tabular}{ccccc}
\hline
 recons loss& PointNet+Max & PointNet +NetVlad & EdgeConv+Max & EdgeConv+NetVlad \\ \hline
No & 86.7 & 87.4 & 90.2 & 90.4 \\ 
Yes & 89.9 & 90.6 & 92.4 & 92.7 \\ \hline
\end{tabular}}
\end{table}

\subsection{Robustness to Noise}
In this section, we evaluate the robustness of our proposed method, 3DCapsule, with respect to two different types of data corruption: 1) outlier points, and 2) point perturbation. This robustness test was carried out in a similar way as the ablation study in Section~\ref{sec:ablation_study}, with the only difference being that the different network combinations were here subjected to varying levels of training and testing noise.

\begin{figure*}
\begin{minipage}[b]{0.2\linewidth}
  \centering
  \centerline{\includegraphics[width=4.9cm]{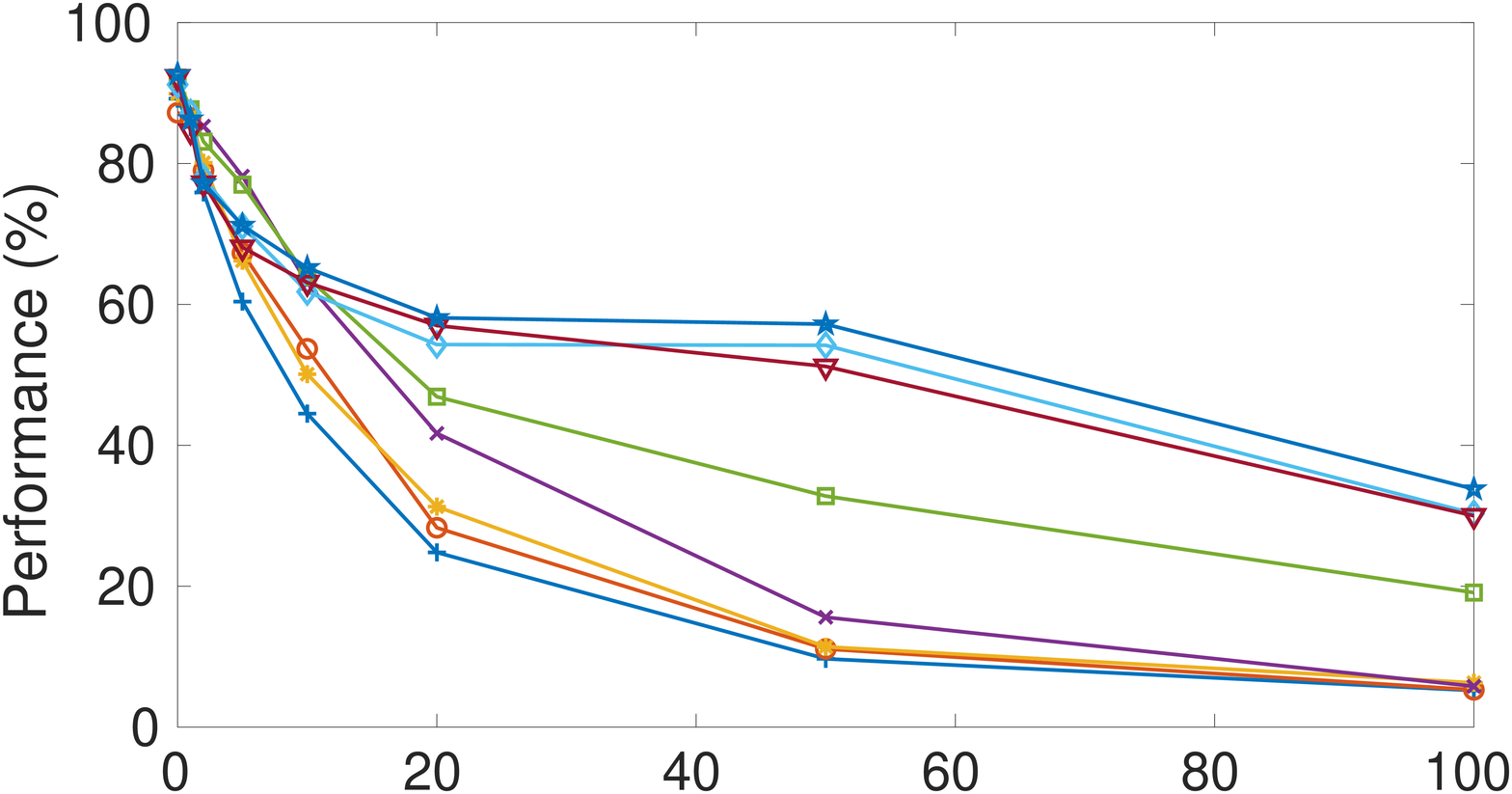}}
  \centerline{Outlier points in training set: 0}\medskip
\end{minipage}
\begin{minipage}[b]{0.2\linewidth}
  \centering
  \centerline{\includegraphics[width=4.9cm]{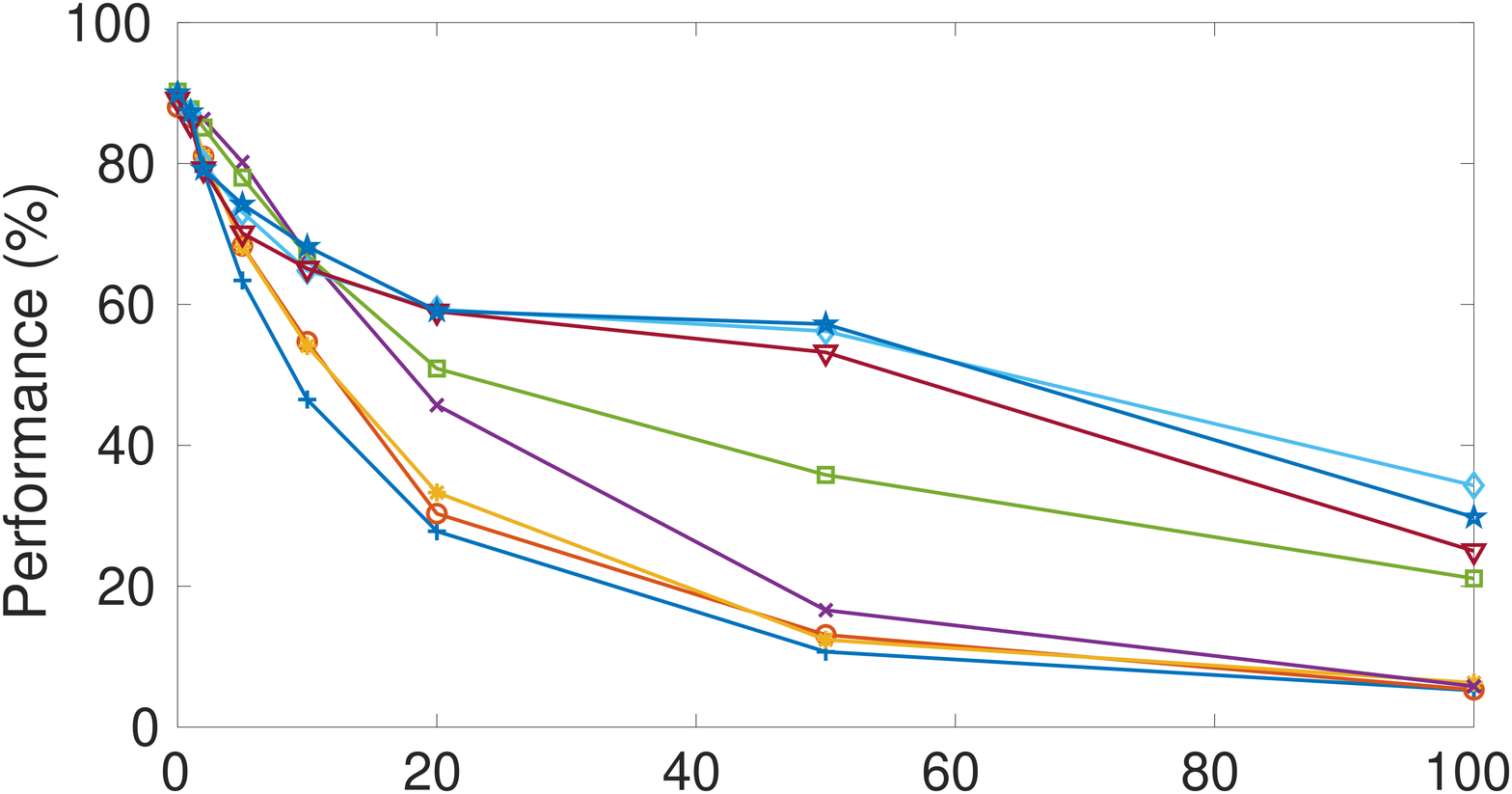}}
  \centerline{Outlier points in training set: 1}\medskip
\end{minipage}
\begin{minipage}[b]{0.2\linewidth}
  \centering
  \centerline{\includegraphics[width=4.9cm]{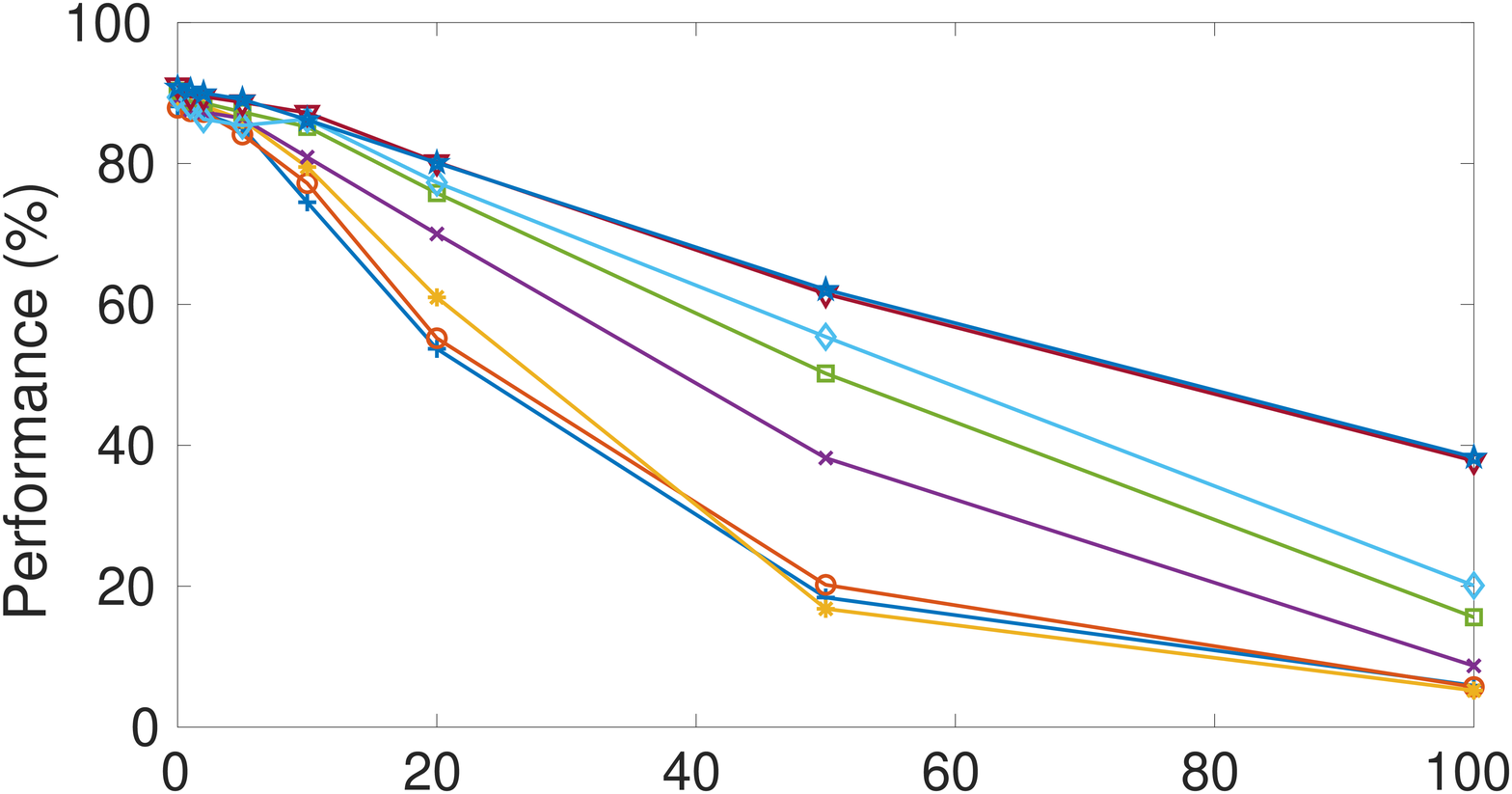}}
  \centerline{Outlier points in training set: 2}\medskip
\end{minipage}
\begin{minipage}[b]{0.2\linewidth}
  \centering
  \centerline{\includegraphics[width=4.9cm]{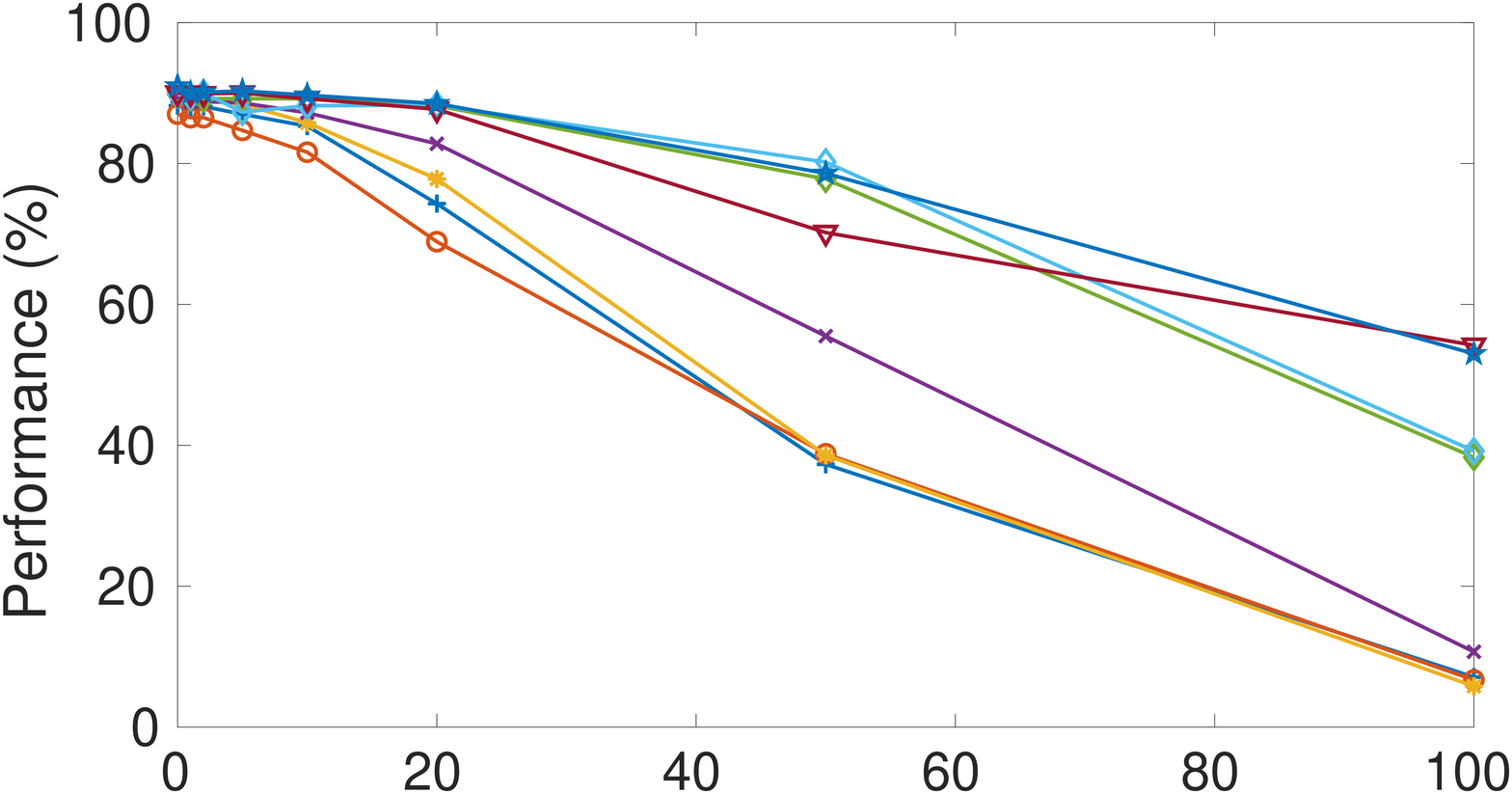}}
  \centerline{Outlier points in training set: 5}\medskip
\end{minipage}
\begin{minipage}[b]{0.2\linewidth}
  \centering
  \centerline{\includegraphics[width=4.9cm]{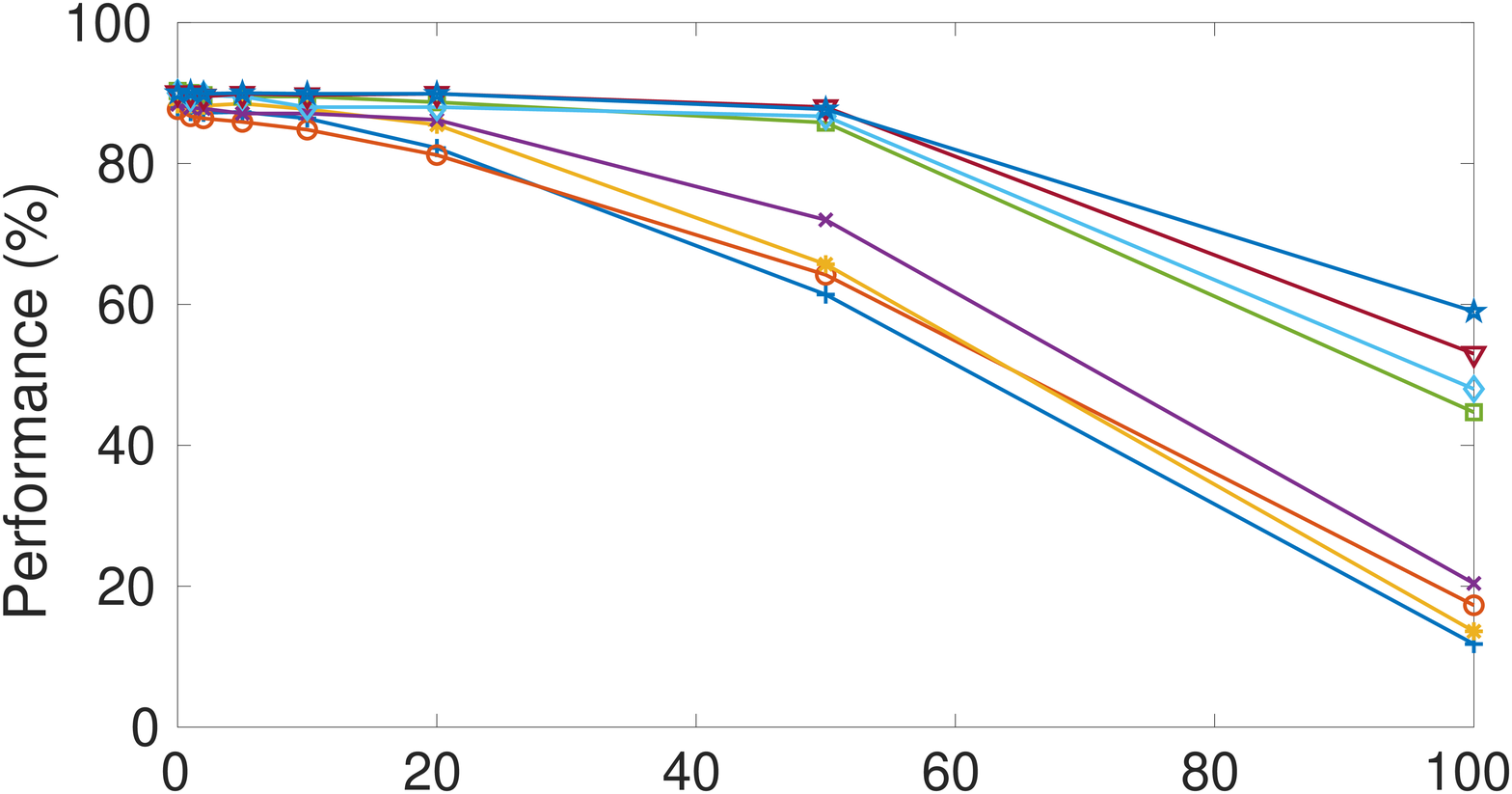}}
  \centerline{Outlier points in training set: 10}\medskip
\end{minipage}
\begin{minipage}[b]{0.2\linewidth}
  \centering
  \centerline{\includegraphics[width=4.9cm]{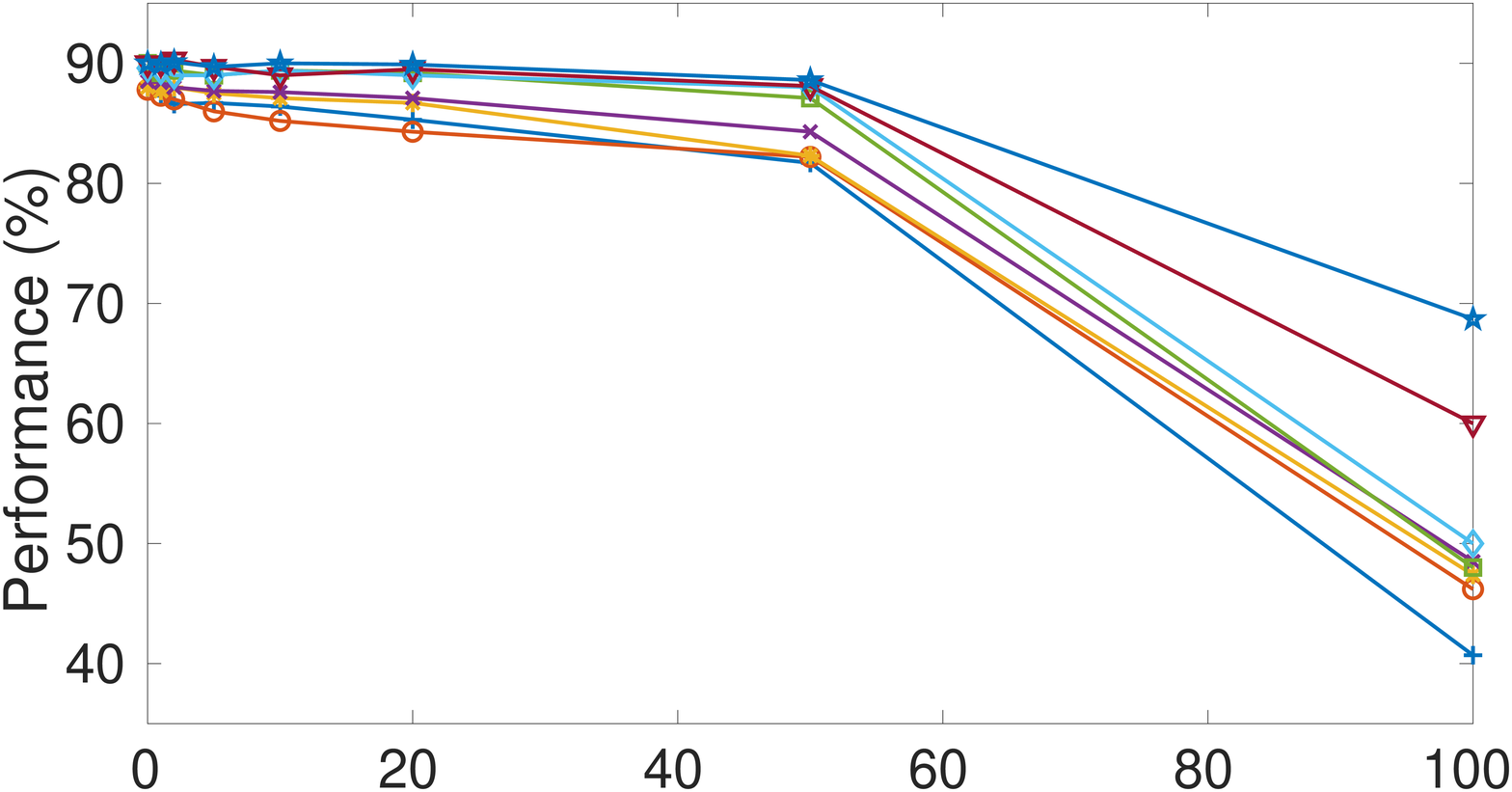}}
  \centerline{Outlier points in training set: 20}\medskip
\end{minipage}
\begin{minipage}[b]{0.2\linewidth}
  \centering
  \centerline{\includegraphics[width=4.9cm]{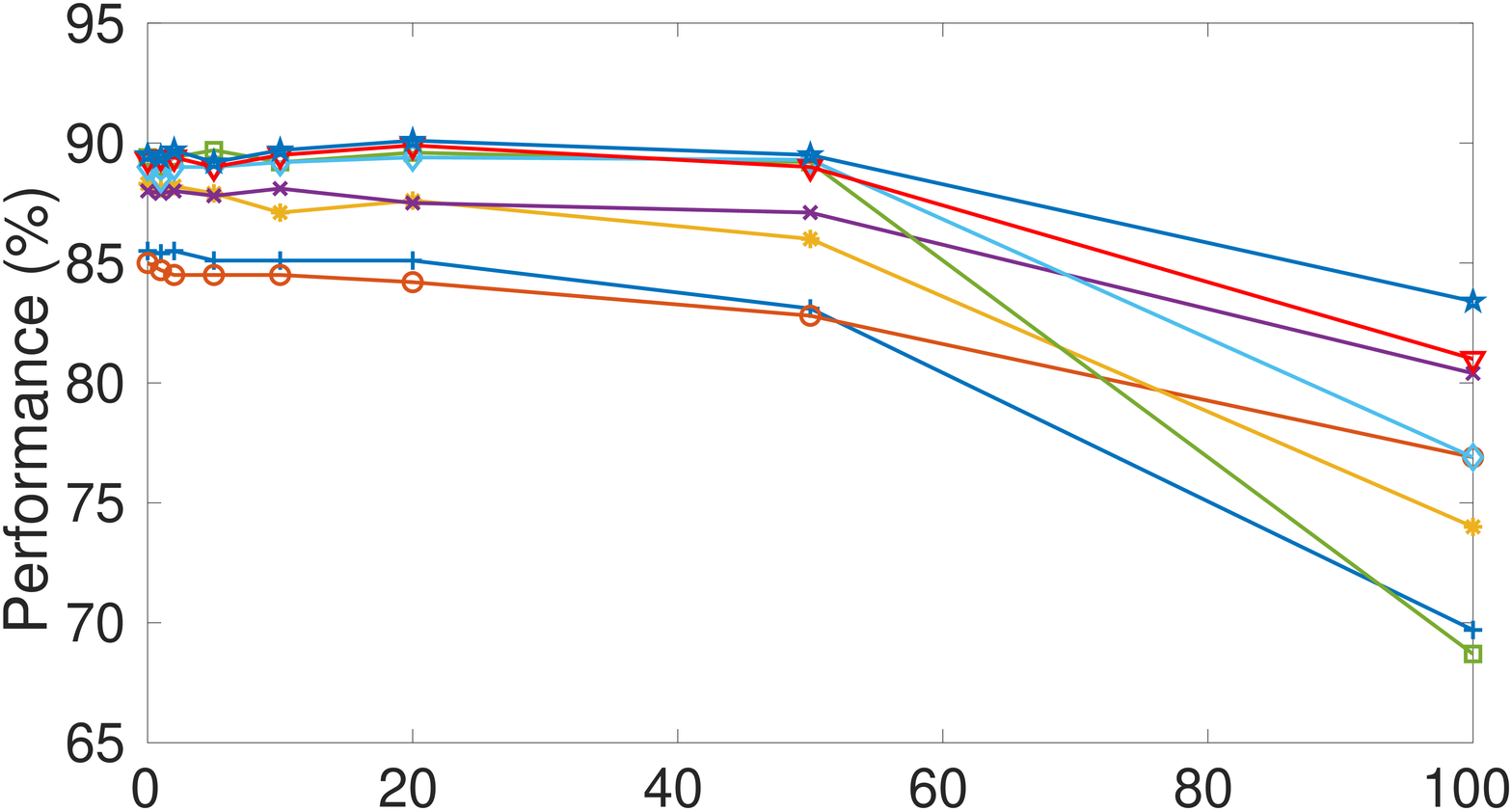}}
  \centerline{Outlier points in training set: 50}\medskip
\end{minipage}
\begin{minipage}[b]{0.2\linewidth}
  \centering
  \centerline{\includegraphics[width=4.9cm]{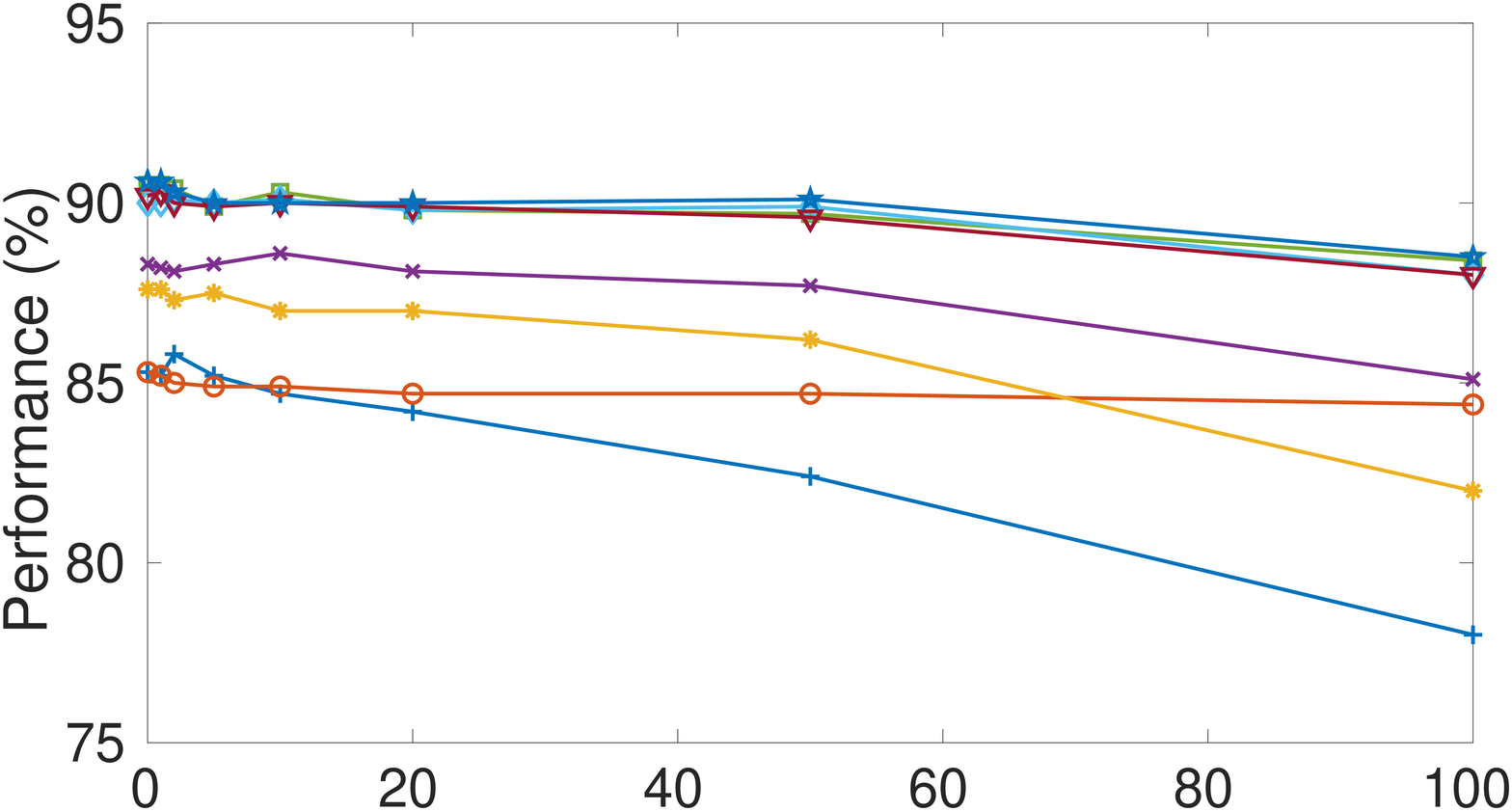}}
  \centerline{Outlier points in training set: 100}\medskip
\end{minipage}
\begin{minipage}[b]{1\linewidth}
  \centering
  \centerline{\includegraphics[width=15.cm]{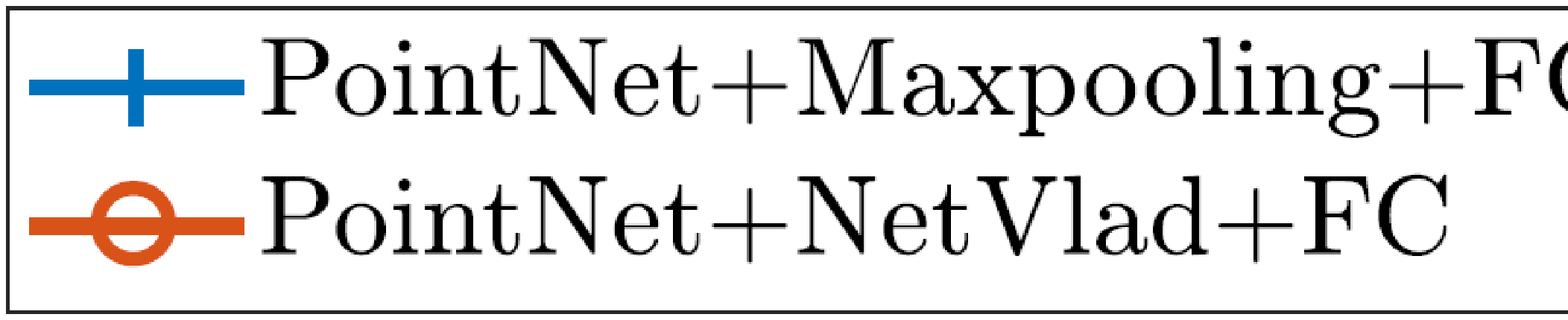}}  
\end{minipage}
\caption{Robustness test with respect to outlier points. The bottom legend applies to all graphs, respective x-axis represents the number of outlier points in the test set, and each graph is an experiment with a different number of outlier points in the training set.}
\label{fig:outlier_test}
\setlength{\abovecaptionskip}{1pt plus 3pt minus 2pt}
\end{figure*}

Input points were normalized to the unit sphere in this experiment. When creating the training data, both distorted and undistorted data were added together, enabling the networks to better learn how to deal with distorted data without unnecessarily reducing performance when exposed to undistorted data.

\noindent
\textbf{Outlier Test:} In the outlier experiment, a varying number of outlier points (0, 1, 2, 5, 10, 20, 50, 100) were replacing the original points, both in the training and the testing samples. The result is shown in Figure~\ref{fig:outlier_test}. It is evident that a 3DCapsule enabled architecture is significantly more robust against outlier points in almost every combination of training and testing noise levels, compared to the baseline (fully connected classifier), irrespectively of which feature extractor and aggregator that is used. Furthermore, it can be seen that NetVlad \cite{Article22} is generally achieving better performance than Maxpooling, and that gap tends to be greater when combined with 3DCapsule as expected.

\begin{figure*}
\begin{minipage}[b]{0.28\linewidth}
  \centering
  \centerline{\includegraphics[width=5.5cm]{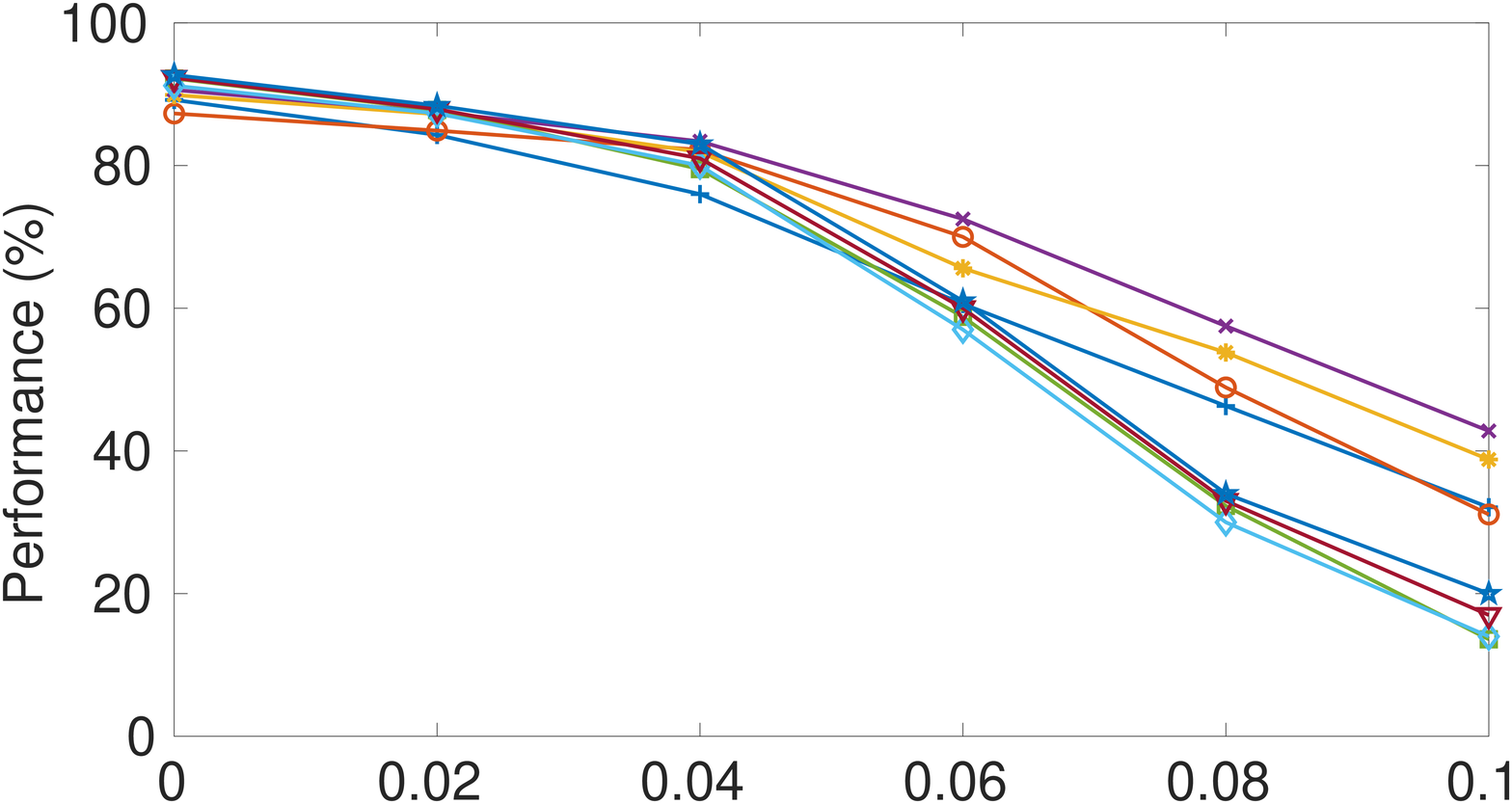}}
  \centerline{Perturbation of the training set: std=0.00}\medskip
\end{minipage}
\begin{minipage}[b]{0.28\linewidth}
  \centering
  \centerline{\includegraphics[width=5.5cm]{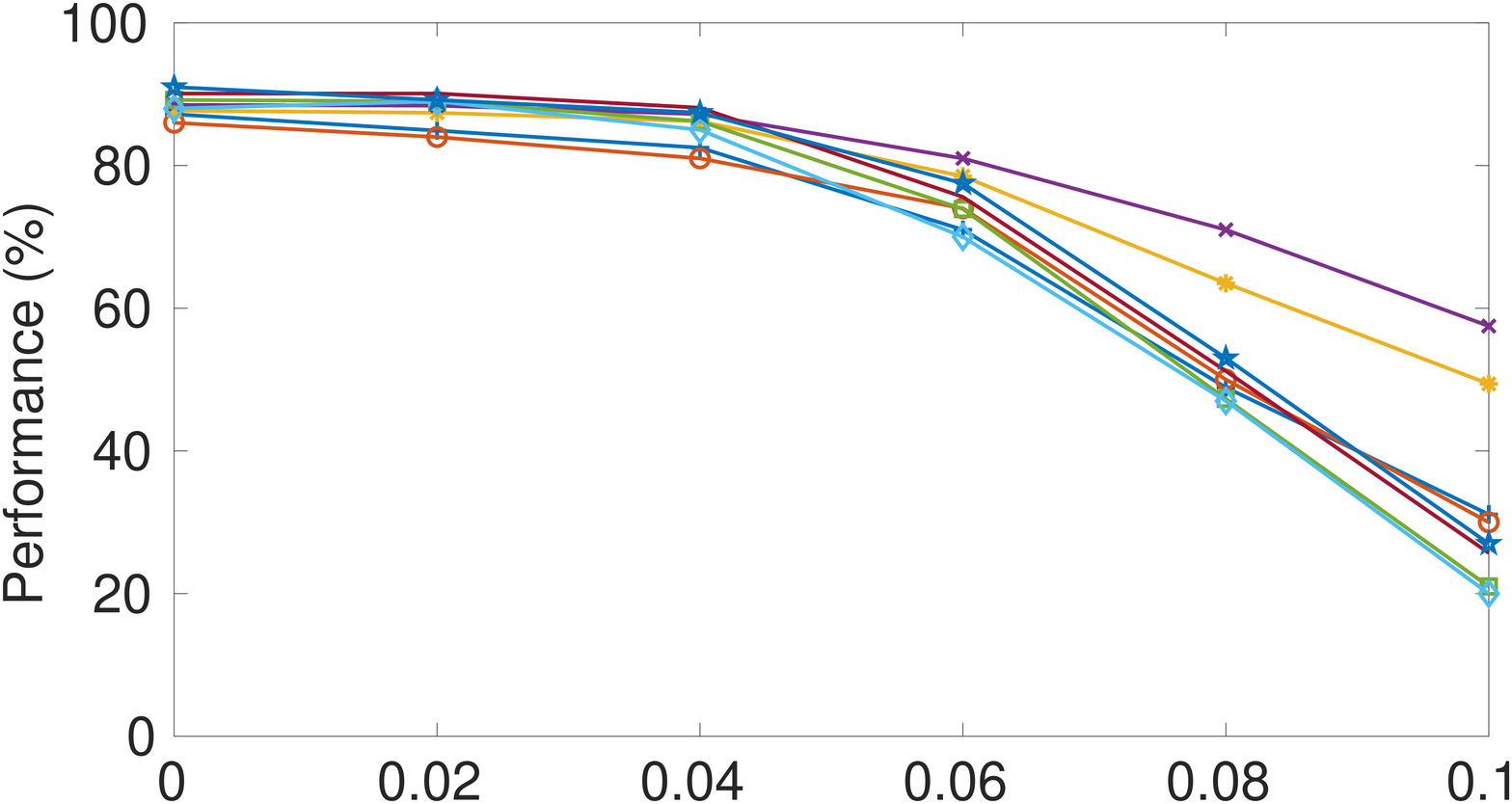}}
  \centerline{Perturbation of the training set: std=0.02}\medskip
\end{minipage}
\begin{minipage}[b]{0.28\linewidth}
  \centering
  \centerline{\includegraphics[width=5.5cm]{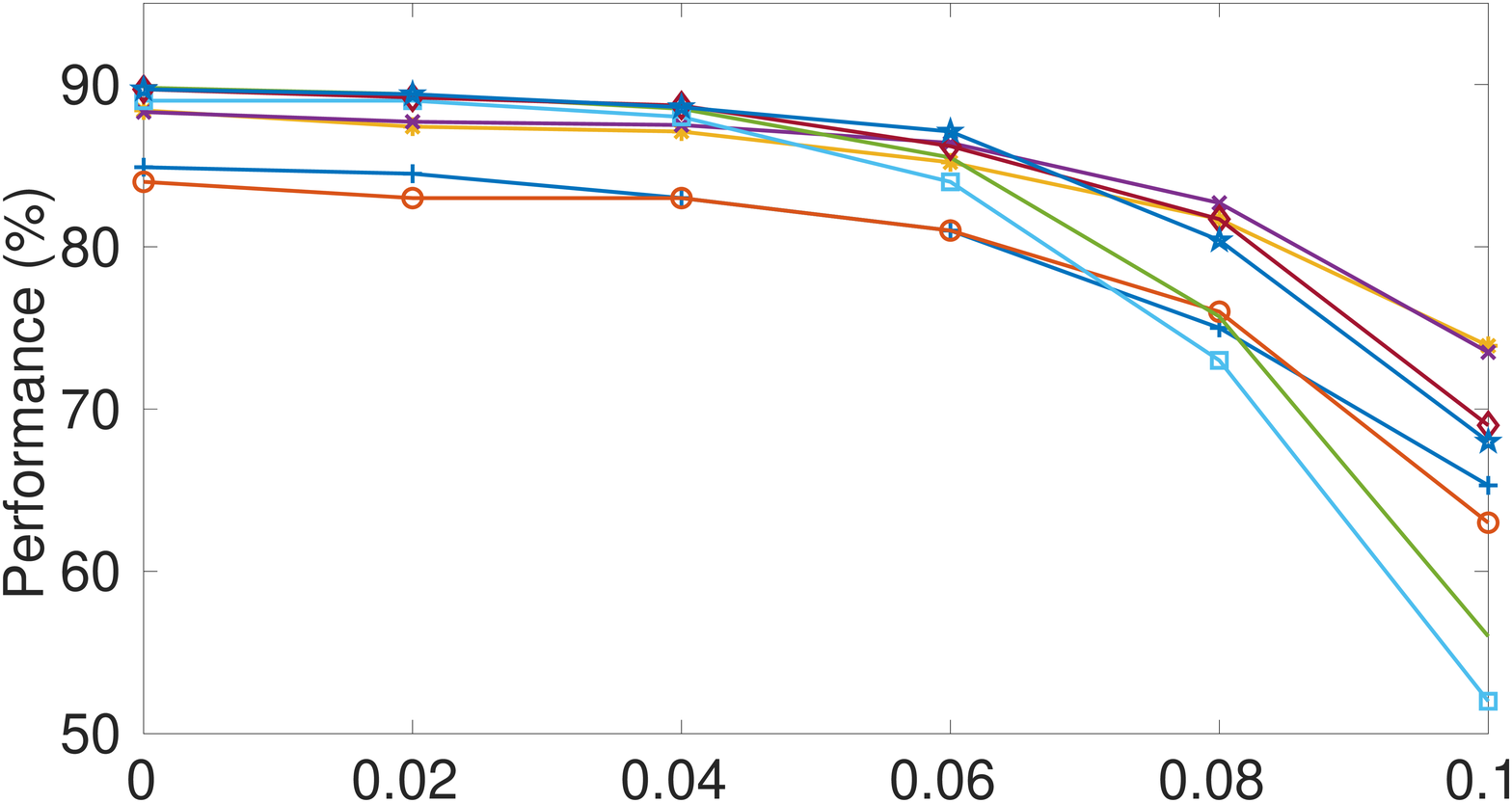}}
  \centerline{Perturbation of the training set: std=0.04}\medskip
\end{minipage}
\begin{minipage}[b]{0.28\linewidth}
  \centering
  \centerline{\includegraphics[width=5.5cm]{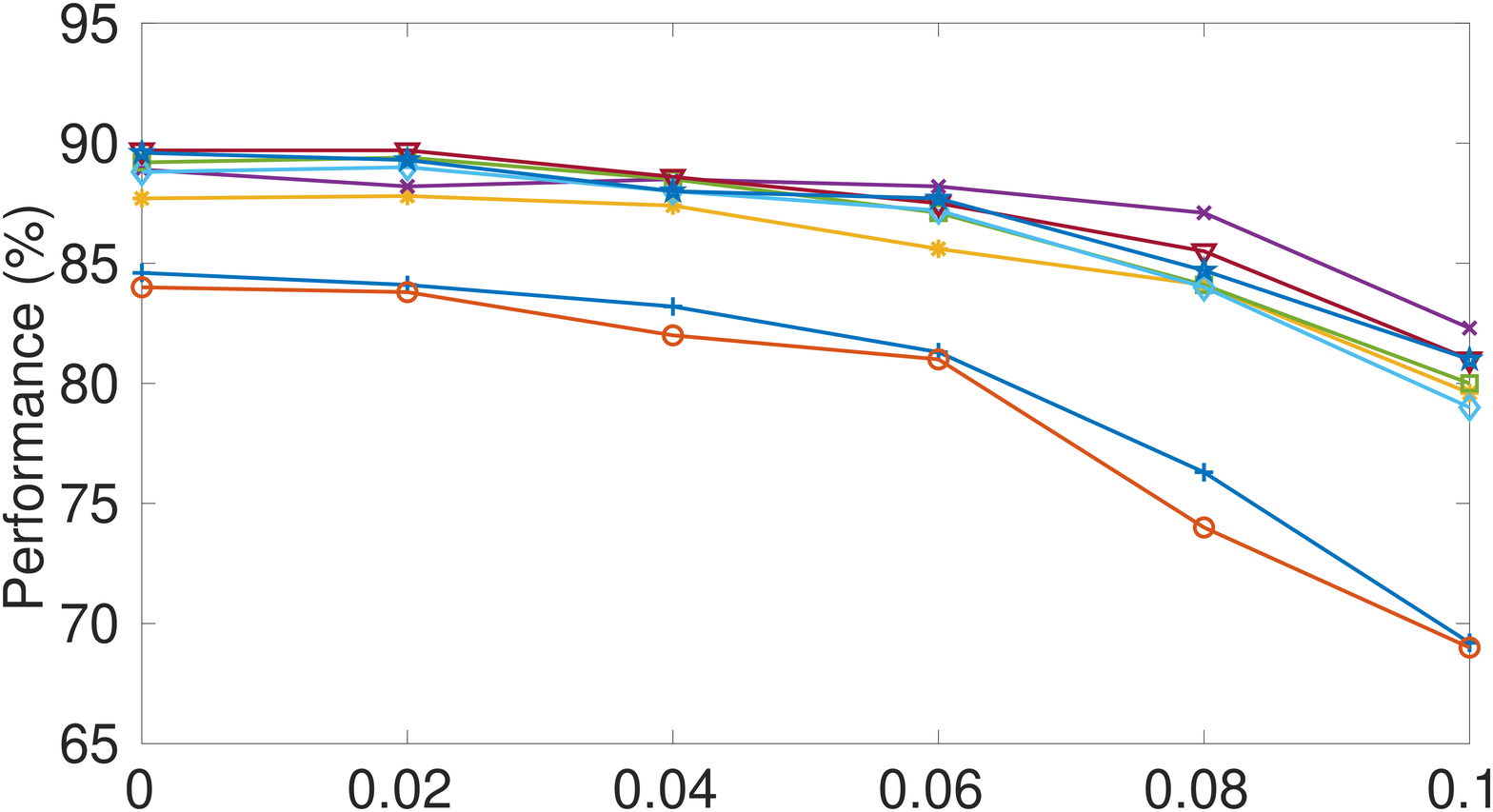}}
  \centerline{Perturbation of the training set: std=0.06}\medskip
\end{minipage}
\begin{minipage}[b]{0.28\linewidth}
  \centering
  \centerline{\includegraphics[width=5.5cm]{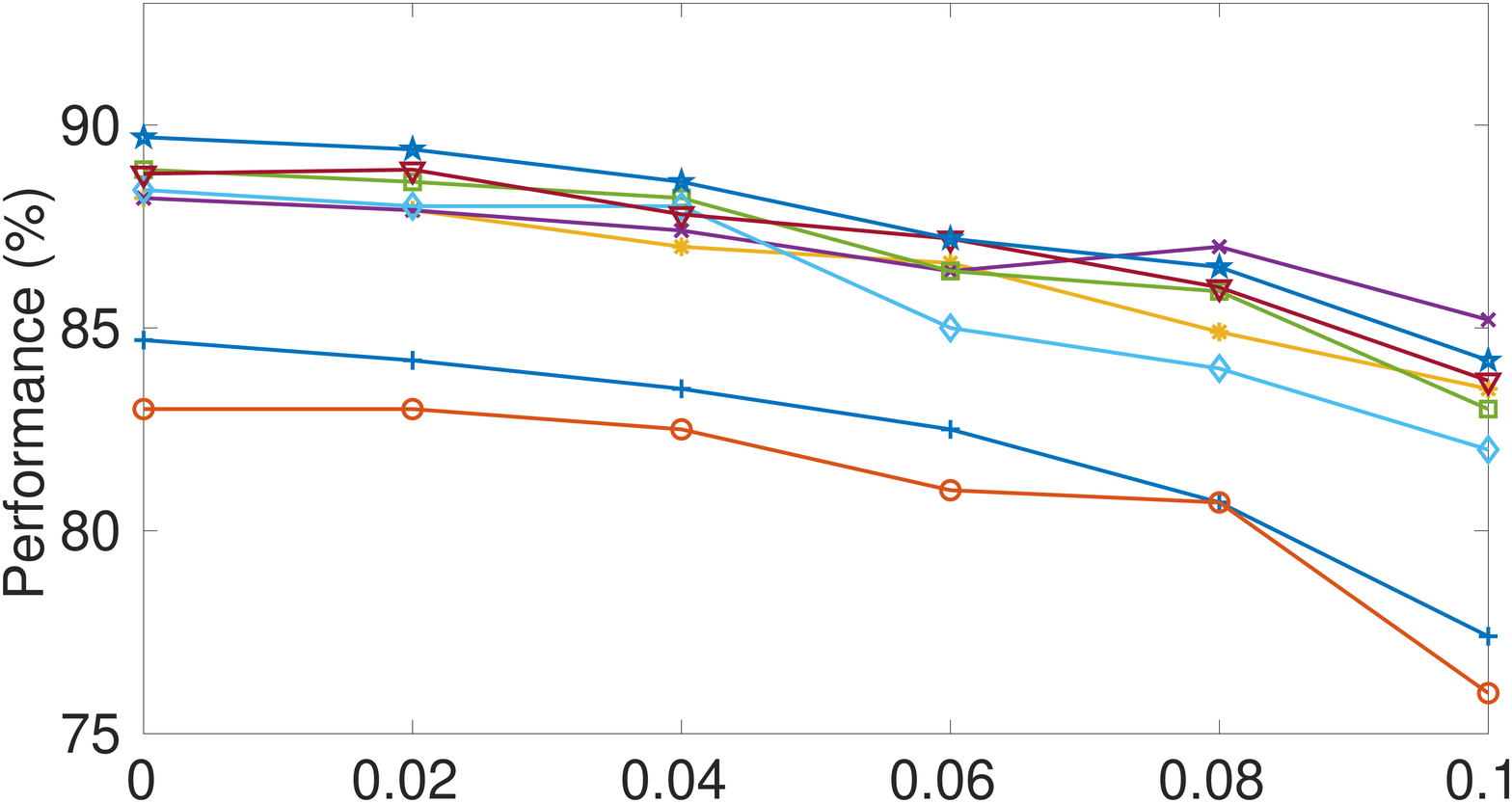}}
  \centerline{Perturbation of the training set: std=0.08}\medskip
\end{minipage}
\begin{minipage}[b]{0.28\linewidth}
  \centering
  \centerline{\includegraphics[width=5.5cm]{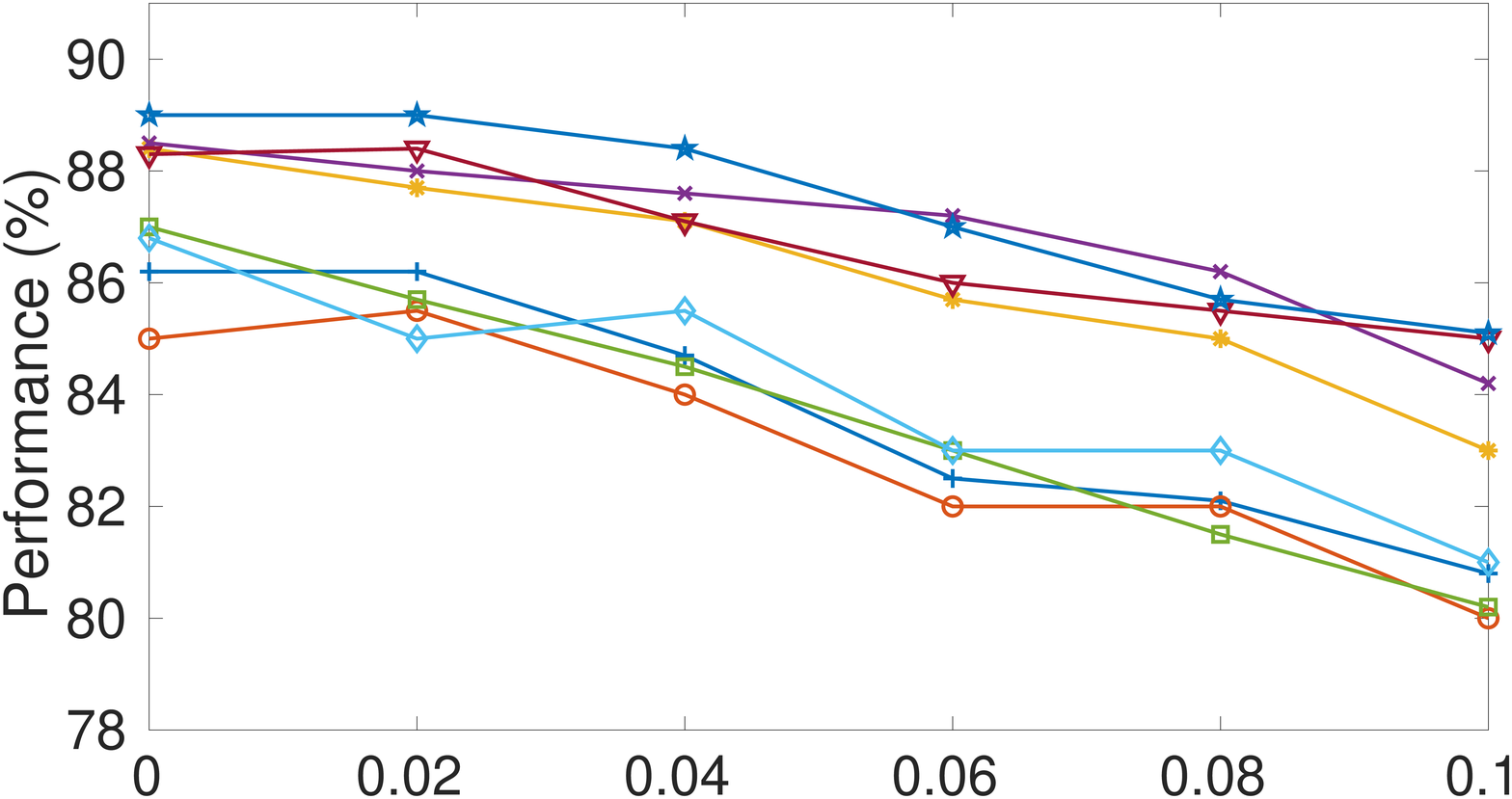}}
  \centerline{Perturbation of the training set: std=0.10}\medskip
\end{minipage}
\begin{minipage}[b]{1\linewidth}
  \centering
  \centerline{\includegraphics[width=15cm]{legend1.eps}}  
\end{minipage}
\caption{Robustness test with respect to perturbed points. The bottom legend applies to all graphs, respective x-axis represents the magnitude of the Gaussian noise in terms of standard deviation that was added to the test set, and each graph is an experiment with a different magnitude of Gaussian noise in the training set.}
\label{fig:perturbation_test}
\setlength{\abovecaptionskip}{1pt plus 3pt minus 2pt}
\end{figure*}

\noindent
\textbf{Perturbation Test:} In the point perturbation test, Gaussian noise was added to each point independently. The standard deviation of the added noise was in the range $[0,0.1]$ in steps of $0, 0.02, 0.04, 0.06, 0.08,$ and $0.1$. The result is shown in Figure~\ref{fig:perturbation_test}. It is evident that, again, a 3DCapsule enabled architecture is significantly more robust, also in the case of Gaussian perturbation noise, compared to the baseline that instead utilises a fully connected classifier. It can also be seen in the graphs that the fully connected classifier is not good at making use of the richer representation that NetVlad \cite{Article22} provides, whereas 3DCapsule can unlock the information provided by the more powerful NetVlad aggregator.

\subsection{Absolute Classification Performance}
In this paper, we do not explicitly set out to construct a new architecture that will achieve state-of-the-art classification performance, but rather contribute with the design of a new component, \textbf{3DCapsule}, that can be drop-in inserted into many other architectures. However, for completeness, we here compare our proposed method to state-of-the-art classification architectures operating on raw, unordered, point clouds that are unaffected by any additional noise.

We chose to compare the best combination of feature extractor and aggregator together with our 3DCapsule that we had found in the experiments in the previous sections. That is, EdgeConv plus NetVlad were used as feature extraction and aggregation modules respectively, together with our 3DCapsule. The comparison was made using the ModelNet40/ModelNet10 3D object recognition benchmark, and the results are shown in Table~\ref{table:state_of_the_art}.

The lower part of the Table shows those methods that are apples-to-apples comparable with ours. That is, they use the same input data representation that we do (3D points only without any additional information such as, e.g., surface normals), and they also use the exact same number of input points, $N=1024$. It is evident that in an apples-to-apples comparison where we have applied our 3DCapsule, we outperform current state-of-the-art. In fact, it is only outperformed by SO-Net \cite{Article27} when they are using significantly more points than we do (5000 vs 1024) as well as using surface normals.

\setlength{\tabcolsep}{4pt}
\begin{table}[t]
\begin{center}
\caption{Classification results on ModelNet40/ModelNet10. The bottom part represents an apples-to-apples comparison with ours using the same type and size of input, and the top part is the case when the restrictions on the input data are relaxed. Also, pc stand for point cloud.}
\vspace{4mm}
\label{table:state_of_the_art}
\scalebox{.75}{

\begin{tabular}{lcccc}
\hline\noalign{\smallskip}
Method &Representation&Input& ModelNet40&ModelNet10\\
\noalign{\smallskip}
\hline
\noalign{\smallskip}

SO-Net\cite{Article27} & pc &$2048\times3$& 90.9&93.9\\
Kd-Net\cite{Article3} &pc & $2^{15} \times3$ & 91.8&93.5\\
SpiderCNN\cite{Article28} & pc + normal &$1024\times6$& 92.4&-\\
Local Spectral\cite{Article29} & pc + normal &$2048\times6$& 92.1&-\\
PointNet++\cite{Article2} & pc + normal&$5000\times6$& 91.9&-
\\

SO-Net\cite{Article27} & pc + normal &$5000\times6$& \bf{93.4}&\bf{95.7}\\

\hline\hline
PointNet\cite{Article1}&pc& $1024\times3$  & 89.2&-\\
PointNet++\cite{Article2} &pc&$1024\times3$&  90.6&-\\
EdgeConv\cite{Article24} & pc &$1024\times3$& 92.2&-\\
Local Spectral\cite{Article29} & pc &$1024\times3$& 91.5&-\\
\hline

\bf{Ours} & pc &$1024\times3$& \bf{92.7}&\bf{94.7}\\
\hline
\end{tabular}}
\end{center}
\end{table}
\setlength{\tabcolsep}{1.4pt}

\subsection{Time and Space Complexity}
It is clear from Table~\ref{table:time} that the 3DCapsule adds additional complexity to the architectures to which it is applied. However, we believe that the benefit of significant increase in robustness to various kinds of noise (as shown in Figure~\ref{fig:outlier_test} and Figure~\ref{fig:perturbation_test}) outweighs these drawbacks. The performance on classification, without added noise, is also increased to a degree, which demonstrates the benefit of 3DCapsule in comparison to the common fully connected classifier, albeit at the cost of an increase in memory usage and inference time.

\begin{table}
\caption{Time and space complexity of point cloud based networks in ModelNet40 classification.}
\vspace{4mm}
\centering
\label{table:time}

\scalebox{.85}{
\begin{tabular}{l c c c}
\hline
Method & Size/MB & Forward/ms & Train/h \\ \hline

PointNet++ \cite{Article2} & 12 & 163.2 & 20 \\ 

SO-Net \cite{Article27} & 11.5 & 59.6 & 1.5 \\ \hline \hline
PointNet \cite{Article1} & 40 & 25.3 & 3-6 \\ 
PointNet \cite{Article1}+3DCapsule & \multicolumn{1}{c}{71} & \multicolumn{1}{c}{116} & \multicolumn{1}{c}{8} \\ \hline \hline
EdgeConv \cite{Article24} & 21 & 94.6 & - \\ 
EdgeConv \cite{Article24}+3DCapsule & \multicolumn{1}{c}{52} & \multicolumn{1}{c}{154} & \multicolumn{1}{c}{12} \\ \hline
\end{tabular}}
\end{table}

\section{Conclusion}
In this paper, we introduced the \textbf{3DCapsule} which is an extension of the Capsule concept that makes it applicable to unordered point sets. It is intended as an alternative, drop-in replacement, in the case of 3D point cloud classification, to the commonly used fully connected classifier. A key insight is that the original Capsule concept, that operated on 2D images, implicitly relied on the fact that the pixels in such images are ordered and has a spatial relationship. Modern architectures for 3D point set classification typically introduce symmetrical functions to achieve invariance to point set permutation, however, this also invalidates the spatial relationship between feature vectors. Therefore, in this work, a \textbf{ComposeCaps} layer was introduced to find a new, meaningful, mapping of the feature vectors to capsules, that the 3DCapsule network could better exploit.

While achieveing improved classification performance in an apples-to-apples comparison against state-of-the-art methods on data that is not affected by additional noise, it is overwhelmingly demonstrated that using the 3DCapsule is superior when subjected to corrupted data such as a varying amount of outliers, or Gaussian noise. This is demonstrated against the typical fully connected classifier, and irrespectively of which feature extractor or aggregator module that it is combined with.

\newpage

{\small
\bibliographystyle{ieee}
\bibliography{egbib}
}

\end{document}